\documentclass[preprint,11pt,authoryear,times]{elsarticle}
\usepackage[a4paper, margin=1in]{geometry}
\usepackage{setspace}
\usepackage{amsmath,amssymb,amsthm}
\usepackage{bm}

\usepackage{graphicx}
\usepackage{booktabs}
\usepackage{makecell}
\usepackage{multirow}
\usepackage{algorithm}
\usepackage{siunitx}
\usepackage{algpseudocode}
\usepackage{tabularx}
\usepackage{xcolor}
\usepackage{rotating}

\newcommand{\red}[1]{\textcolor{red}{#1}}
\usepackage{array}
\usepackage{threeparttable}
\usepackage{wrapfig}
\usepackage{siunitx}
\usepackage{tikz}
\usetikzlibrary{arrows.meta,decorations.markings}

\usepackage{xcolor}
\definecolor{mygray}{RGB}{245,245,245}
\definecolor{myecru}{RGB}{243,241,222}
\definecolor{myteal}{RGB}{230,241,253}
\definecolor{myblue}{RGB}{31,74,170}
\definecolor{boundary}{RGB}{70,69,68}

\newcolumntype{Y}{>{\raggedright\arraybackslash}X}
\newcolumntype{P}[1]{>{\raggedright\arraybackslash}p{#1}}

\usepackage[hidelinks]{hyperref}
\usepackage[capitalise,nameinlink]{cleveref}

\usepackage{comment}

\journal{European Journal of Operational Research}

\begin{document}

\begin{frontmatter}

\title{Decision-Value Attribution in Predict-then-Optimize Systems}

\author[au]{Konstantinos Ziliaskopoulos}
\ead{kzz0034@auburn.edu}

\author[au]{Alexander Vinel\corref{cor1}}
\ead{alexander.vinel@auburn.edu}

\author[ua]{Alice E. Smith}
\ead{alice.smith@ua.edu}

\cortext[cor1]{Corresponding author}

\affiliation[au]{
  organization={Auburn University},
  addressline={Industrial \& Systems Engineering, 36849},
  city={Auburn},
  state={AL},
  country={USA}
}

\affiliation[ua]{
  organization={University of Alabama},
  addressline={Mechanical Engineering \& Computer Science, 35487},
  city={Tuscaloosa},
  state={AL},
  country={USA}
}

\begin{abstract}
Predictive models are increasingly embedded in operational decision-making, yet standard explanation methods typically explain forecasts rather than the decisions those forecasts induce. This distinction is important in predict-then-optimize systems: large forecast changes may leave the optimizer's action unchanged, while small changes can alter the selected decision and its realized value. We propose Decision-Value Attribution (DVA), a Shapley-based framework for attributing the value of a fixed prediction--optimization pipeline. The framework defines cooperative games whose payoff is the downstream decision value, allowing the players to be information sources, optimization or design parameters, or both. We present three variants: InfoDVA attributes value to features, DesignDVA attributes value to operational configurations, and Decision-Value Interactions (DVI) quantifies how information and design jointly create value. We further distinguish post-DVA, which evaluates decisions using realized outcomes, from pre-DVA, which evaluates decisions under the model's full prediction. This separation turns attribution into a decision-level diagnostic of whether the model's operational beliefs align with realized performance. The resulting attributions are expressed in the units of the operational objective and decompose the gain or loss relative to a baseline. Case studies in electricity storage arbitrage and emergency medical service coverage show that predictive explanations can be poor proxies for operational value, that DVA can guide targeted information-control interventions, and that optimization configurations determine when predictive information is decision-relevant.
\end{abstract}


\end{frontmatter}

\section{Introduction}
\begin{wrapfigure}{r}{0.4\linewidth}
    \centering
    \vspace{-1em}
    \includegraphics[
        width=0.46\linewidth,
        trim=800pt 200pt 800pt 20pt,
        clip]{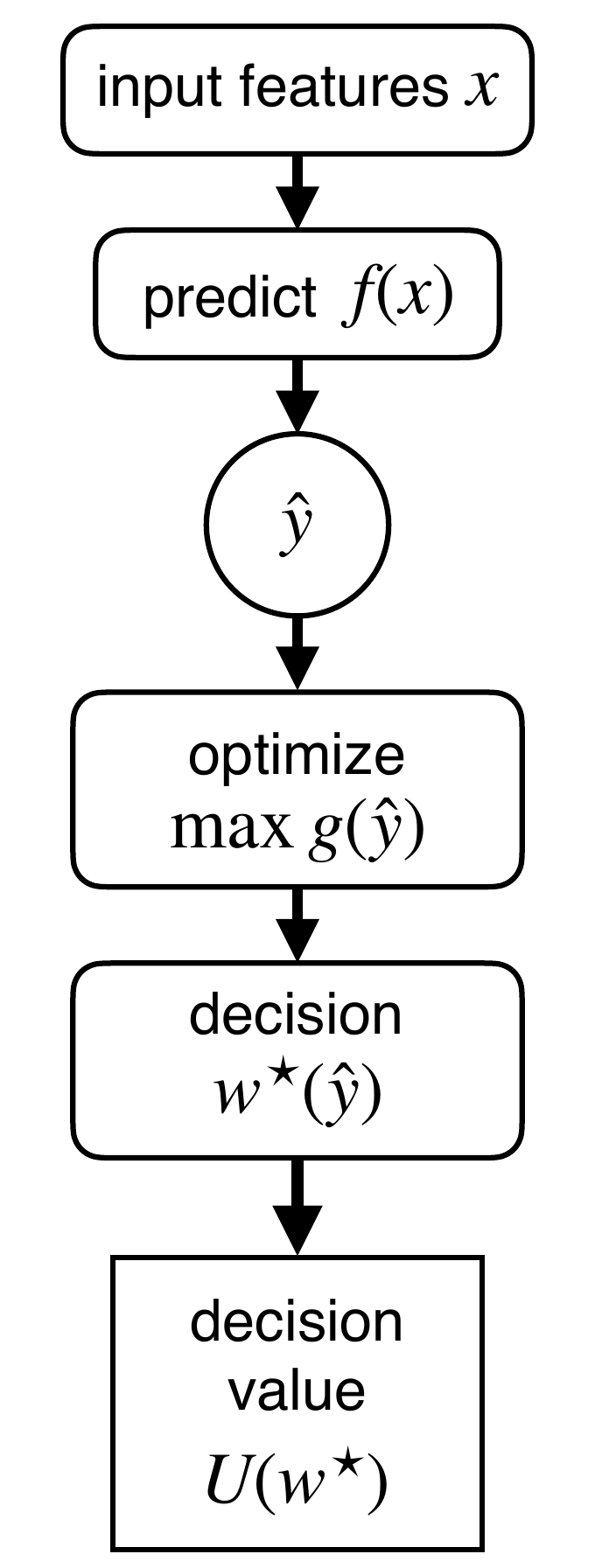}
    \caption{Overview of the Predict-then-Optimize (PtO) pipeline.}
    \label{fig:pto-pipeline}
    \vspace{-2em}
\end{wrapfigure}

In many applications a critical prediction is not the final object of interest. For example, a battery operator forecasts electricity prices to schedule charging and discharging, a logistics planner forecasts travel times to select routes, or a retailer forecasts demand to set inventory levels or assortment decisions. These are predict-then-optimize (PtO) systems, as shown in \cref{fig:pto-pipeline} where a predictive model estimates uncertain parameters and an optimization model selects an optimal decision based on the upstream model's prediction  \citep{bertsimas2020predictive,sadana2025survey}. In these systems, the value of information is primarily determined by how much it improves the decision induced by that prediction.

This can create a mismatch for standard explanation methods. Popular feature attribution tools such as LIME and SHAP explain the mapping from input features to model output \citep{ribeiro2016should,lundberg2017unified}. This is useful for auditing the forecasting model, but it does not necessarily explain the PtO system. A feature may substantially change a prediction while leaving the optimizer's selected decision unchanged. Conversely, a small predictive change can be operationally decisive if it moves the prediction across a decision boundary and changes the selected action. In operational settings, therefore, predictive importance and decision importance can diverge.
We devise \emph{Decision-Value Attribution} (DVA), a Shapley-based framework for attributing the value of a PtO pipeline to the components that generate, or shape, the downstream decision. The main idea, following the Shapley method, is to define a cooperative game whose payoff is downstream decision value. The players may be information sources, design parameters, or both. The resulting attribution decomposes the gain or loss from using a given information set or design configuration, expressed directly in the units of the operational objective.

The first component of this approach is \emph{Information DVA} (InfoDVA), where the players are features or feature groups. InfoDVA answers the question of which information sources improve or degrade the value of the decision induced by the PtO pipeline. The second component is \emph{DesignDVA}, where the players are design or optimization parameters such as storage capacity, power limits, fleet size, or coverage radius. DesignDVA is a form of sensitivity analysis that allows comparison of configurations against a chosen reference. Its main role is to support \emph{JointDVA}, where both information sources and design parameters are players in the same cooperative game. JointDVA answers the question of how the decision value is jointly created by the data available to the predictive model and by the parameters of the optimization model.
We then introduce \emph{Decision-Value Interactions} (DVI), which extend interaction attribution methods to DVA games using Faith-Shap interaction indices \citep{tsai2023faith}. DVI measures whether groups of players create decision value nonlinearly, meaning the value function over player contributions cannot be expressed as a linear expression of individual effects, as PtO, by design, requires consideration of the interaction between information and design. For example, for an energy system, weather information may be valuable only when a battery has sufficient capacity to exploit temporal price arbitrage, or spatial demand information may matter only when a facility location budget makes alternative deployments feasible. These feature--parameter interactions are difficult to detect with predictive explanations alone.

For each of these games, we define two evaluation modes. \emph{Post-DVA} evaluates the decision using the realized outcome and is therefore an ex-post diagnostic of realized operational value. \emph{Pre-DVA} replaces the realized outcome, or actual observation, with the full model prediction and asks how a feature affects decision value from the model's own point of view. Pre-DVA is useful before realized outcomes are observed and thus can be used during inference, while post-DVA is useful for auditing the performance of PtO pipelines and can identify useful interventions.

This paper makes three contributions. First, we adapt Shapley value attribution to predict-then-optimize systems, attributing downstream decision value rather than predictive output, and develop this into the Decision Value Attribution (DVA) framework. DVA unifies information, design, and joint attributions and separates model-implied from realized usefulness.  Second, we extend this framework to Decision-Value Interactions (DVI), which identify complementarities and substitutes among information sources and optimization parameters. Third, we demonstrate the framework in two case studies, that of electricity storage arbitrage and emergency medical services (EMS) coverage.  We show that predictive explanations can be poor proxies for operational value, and that DVA can guide targeted information-control interventions and that optimization design determines when information is decision-relevant. To support replication and reuse, we share all necessary implementations.\footnote{All code and data used in this paper are publicly available at \href{https://github.com/kziliask/DVA}{\texttt{github.com/kziliask/DVA}}.}

The remainder of the paper is organized as follows. Section~\ref{sec:related-work} positions the paper within prescriptive analytics, explainable AI, explainable optimization, and value-of-information research. Section~\ref{sec:methods} presents the DVA framework. \cref{sec:experiments} presents the two case studies which demonstrate our method. Finally, Section~\ref{sec:discussion} discusses managerial implications and limitations, and concludes.

\section{Related Work}
\label{sec:related-work}

\paragraph{Predict-then-optimize and prescriptive analytics}

This paper is related to the areas of prescriptive analytics and contextual optimization, where data are used to improve decisions under uncertainty. The main takeaway of this literature is that predictive models should be evaluated through the decisions they support rather than through prediction error alone \citep{bertsimas2020predictive}. Therefore, decision-focused learning and PtO methods develop training procedures that account for downstream optimization performance \citep{donti2017task,mandi2024decision}. Related work also shows that in optimization problems with structured feasible sets, downstream performance may depend more on the ranking of feasible decisions than on pointwise accuracy \citep{mandi2022decision}. Our paper addresses a different question. Instead of aiming to propose a new training method or an optimizer, we take a trained predictive model, an optimization configuration, and a decision rule, and ask how the value of the resulting decision should be attributed. As a result, DVA complements decision-focused learning in that it explains which information sources and design choices contribute to the operational value of its decisions, once processed by the preferred PtO approach.

\paragraph{Shapley explanations and predictive feature attribution}

An established body of literature studies post-hoc explanations of black-box predictive models. Of the two most popular approaches, LIME constructs local surrogate models around a prediction \citep{ribeiro2016should}, while SHAP uses Shapley values to attribute a model output to input features \citep{lundberg2017unified,shapley1953value}. Extensions of Shapley-based explanations address feature dependence, alternative value functions, local and global explanations, and interaction effects \citep{aas2021explaining,borgonovo2024many}. This paper builds on the insight that Shapley explanations are defined by a value function. Existing Shapley-based methods typically choose value functions that explain the predictive model itself, which describe how input features affect the model output. In contrast, DVA chooses a value function defined on the downstream decision problem. It therefore attributes changes in the value generated by the decision that uses the prediction. This shift is important because prediction errors and decision consequences need not align. Large prediction changes may have no operational effect if they leave the optimal decision unchanged, while small changes near a decision boundary may substantially change decision value.

\paragraph{Explainable optimization and explainable OR} A growing stream of work studies explainability for optimization and operations research. Recent work on explainable AI for operational research emphasizes the need to connect predictive models, optimization models, and decision support systems \citep{de2024explainable}. Other papers explain data-driven optimization decisions through counterfactual contexts, interpretable surrogate models, or feature-based optimization models \citep{kurtz2025counterfactual,forel2023explainable,goerigk2023framework,goerigk2024feature}. These approaches are valuable when the goal is to explain an optimization decision and construct an interpretable decision rule. DVA, however, assumes fixed predictive and downstream optimization models, and the analyst aims to attribute the value of the decision returned by this pipeline. The attribution is therefore pipeline-aware. A change in the predictive model, the optimization formulation, the reference design, or the evaluation mode can alter the DVA game and affect the resulting explanation.

\paragraph{Interactions, value of information, and data valuation} DVA is also connected to value-of-information and data-valuation ideas. Classical value-of-information theory evaluates information by its effect on decisions \citep{howard1966information}. Data valuation methods, such as Data Shapley, allocate model-performance value to training observations \citep{ghorbani2019data}. In our analysis, in contrast, DVA attributes the value of an implemented PtO decision to information sources and design parameters for a particular decision instance or evaluation set. Interaction attribution (across features, design parameters, and feature-design pairs) is important for this purpose. Shapley interaction methods, including the Shapley--Taylor index and Faith-Shap, extend feature attribution from individual players to groups of players \citep{sundararajan2020shapley,tsai2023faith}. We apply these ideas to decision-value games in our approach.

\section{Decision-Value Attribution}
\label{sec:methods}

\subsection{Prediction--optimization pipeline}
Consider a PtO system with an input vector \(x\in\mathcal X\), a trained predictive model
    $f:\mathcal X \rightarrow \Theta,$
and a downstream optimization model. The prediction $\hat y_N = f(x)$
is a vector of estimated problem parameters. The subscript \(N\) denotes the full information set. Let \(\rho\in \mathcal D\) denote an operational design or optimization configuration. Examples include the energy capacity of a battery, the coverage radius of a field operator, or the chosen heuristic. Given a prediction \(\hat y\) and design $\rho$, the optimizer or decision rule returns
\begin{equation}
\label{eq:decision-map}
    g_\rho(\hat y)
    \in
    \arg\max_{w\in\mathcal W(\rho)}
    H(w,\hat y; \rho),
\end{equation}
where \(\mathcal W(\rho)\) is the feasible set and \(H\) is the chosen optimization objective. Assume that ties are resolved by a fixed rule, i.e., \(g_\rho\) is deterministic. If the downstream problem is written as a cost minimization problem, it can be expressed in the form above by taking \(H\) to be a negative cost.
To streamline the presentation, we will denote as \(\tilde y_m\) the true realized outcome in the post-DVA setting, and the full model prediction in the pre-DVA setting.
This allows us to use the same DVA construction to conduct either an ex-post diagnostic or an ex-ante model belief evaluation. Recall that Post-DVA answers, which model components improve or degrade the realized decision value, while Pre-DVA answers, which model components improve the decisions according to the model’s predicted outcome.

\subsection{Coalition predictions}

Let \( N=\{1,\ldots,p\}\) denote the set of information players, usually input features or feature groups. For a coalition \(S\subseteq N\), the coalition prediction \(\hat y_S(x)\) is the prediction obtained when only the features in \(S\) are available. In our case studies we use empirical background marginalization. Given a background set \(B\), define \(-S = N \setminus S\). For any \(x,b \in \mathcal X\), let \(x_S\) denote the hybrid input vector that keeps the components of \(x\) corresponding to players in \(S\) and fills all remaining components using the background observation \(b\). We define
\begin{equation}
\label{eq:coalition-prediction}
    \hat y_S(x)
    =
    \frac{1}{|B|}
    \sum_{b\in B}
    f(x_S,b_{-S}).
\end{equation}
Thus $\hat y_\emptyset = \mathbb E_{x\sim B}[f(x)]$ is the model's average prediction over the background set and \(\hat y_N=f(x)\) is the full information prediction. Other coalition constructions, including conditional versions that account for feature dependence, can be substituted without changing the DVA framework. We show a high-level overview of the games we define, the players they contain, their value functions and a description of what they attribute in \cref{tab:dva-games}.  These are discussed in more detail in the sections that follow.

\begin{table}[t]
\centering
\small
\begin{threeparttable}
\caption{Decision-value attribution games and characteristic functions.}
\label{tab:dva-games}
\begin{tabularx}{\linewidth}{P{0.16\linewidth}P{0.24\linewidth}Y}
\toprule
\textbf{Game} 
& \textbf{Players} 
& \textbf{Characteristic function\tnote{\dag}} \\
\midrule

\textbf{InfoDVA}
& Information players \(S\subseteq N\)
&
\(\displaystyle
v^{\mathrm{Info}}_m(S)
=
\tilde y_m^\top w_S
-
\tilde y_m^\top w_\emptyset,
\qquad
w_S=g(\hat y_S).
\)
\newline Attributes decision value to features or feature groups. \\

\addlinespace

\textbf{DesignDVA}
& Design players \(T\subseteq D\)
&
\(\displaystyle
v^{\mathrm{Design}}_m(T)
=
\tilde y_m^\top w_T
-
\tilde y_m^\top w_\emptyset,
\qquad
w_T=g_{\rho(T)}(\hat y_N).
\)
\newline Attributes decision value to design or optimization parameters. \\

\addlinespace

\textbf{JointDVA}
& Information and design players \(S \subseteq N,T\subseteq D\)
&
\(\displaystyle
v^{\mathrm{Joint}}_m(S,T)
=
\tilde y_m^\top w_{ST}
-
\tilde y_m^\top w_\emptyset,
\qquad
w_{S T}=g_{\rho(T)}(\hat y_S).
\)
\newline Attributes value jointly to information and design components. \\

\addlinespace

\textbf{DVI}
& Player groups \(G\subseteq N\cup D\), \(|G|\geq 2\)
&
\(\displaystyle
\Psi^m(G)
=
\mathrm{FaithShap}(v_m,G),
\)
\newline where \(v_m\) is any of InfoDVA, DesignDVA, or JointDVA games.
Attributes interaction value among DVA players. \\

\bottomrule
\end{tabularx}

\begin{tablenotes}[flushleft]
\footnotesize
\item[\dag] For modes $m \in \{\mathrm{pre, post}\} \  \tilde{y}_m$ represents either the model prediction $\tilde{y}_\mathrm{pre} = \hat y_N  $ or the ground truth observation $ \ \tilde{y}_{\mathrm{post}} = y  $.

\end{tablenotes}
\end{threeparttable}
\end{table}

\subsection{InfoDVA}
\label{subsec:infodva}
InfoDVA attributes decision value to information players while holding the operational design fixed. Fix a configuration \(\rho\). For an evaluation mode \(m\in\{\mathrm{post},\mathrm{pre}\}\), define the InfoDVA game as
\begin{equation}
\label{eq:infodva-game}
    v^{\mathrm{Info}}_{m}(S;\rho)
    =
    U\!\left(g_\rho(\hat y_S),\rho;\tilde y_m\right)
    -
    U\!\left(g_\rho(\hat y_{\emptyset}),\rho;\tilde y_m\right),
    \qquad
    S\subseteq N .
\end{equation}
The null coalition corresponds to the baseline prediction $\hat y_\emptyset = \mathbb E_{x\sim B}[f(x)]$ which is the model's average prediction over the background dataset $B$, and the baseline decision is the decision that $\hat y_\emptyset$ induces. Thus $v^{\mathrm{Info}}_{m}(S;\rho)$ measures the decision value gain or loss from using the coalition prediction \(\hat y_S\) over the baseline prediction.

\begin{figure}[ht]
    \centering
    \resizebox{0.6\linewidth}{!}{%
        \begin{tikzpicture}[
    thick,
    >=Stealth,
    every node/.style={font=\normalsize},
    ylab/.style={font=\large}
]

\definecolor{coneone}{HTML}{EAF4D3}   
\definecolor{conetwo}{HTML}{D9EEF7}   
\definecolor{conethree}{HTML}{E8DDF2} 
\definecolor{myblue}{RGB}{50,90,220}

\def\xmin{-5.2}
\def\xmax{5.8}
\def\ymin{-5.0}
\def\ymax{5.0}

%
\coordinate (O)  at (0,0);
\def\s{1.6}  
\coordinate (W1) at (-1*\s,  2*\s);
\coordinate (W2) at ( 2*\s,  0);
\coordinate (W3) at (-1*\s, -2*\s);

\coordinate (L) at (\xmin,0);
\coordinate (U) at ({\ymax/1.5},\ymax);   
\coordinate (D) at ({-\ymin/1.5},\ymin);  

\fill[coneone, opacity=0.75]
    (O) -- (U) -- (\xmax,\ymax) -- (\xmin,\ymax) -- (L) -- cycle;

\fill[conetwo, opacity=0.75]
    (O) -- (U) -- (\xmax,\ymax) -- (\xmax,\ymin) -- (D) -- cycle;

\fill[conethree, opacity=0.75]
    (O) -- (L) -- (\xmin,\ymin) -- (\xmax,\ymin) -- (D) -- cycle;

\draw[->, line width=1.1pt] (\xmin,0) -- (\xmax+0.35,0) node[below] {$\hat{y}_1$};
\draw[->, line width=1.1pt] (0,\ymin) -- (0,\ymax+0.35) node[right] {$\hat{y}_2$};

\draw[line width=1.7pt] (L) -- (O) -- (U);
\draw[line width=1.7pt] (O) -- (D);

\fill (O) circle (2.7pt);

\fill[white, opacity=0.0] (W1) -- (W2) -- (W3) -- cycle;
\draw[line width=2.5pt] (W1) -- (W2) -- (W3) -- cycle;

\fill (W1) circle (2.6pt);
\fill (W2) circle (2.6pt);
\fill (W3) circle (2.6pt);

\node[above left=1pt] at (W1) {$w_1$};
\node[above right=2pt]      at (W2) {$w_2$};
\node[below left=1pt] at (W3) {$w_3$};

\node at (-4.2, 4.6) {$C(\mathbf{w}_1)$};
\node at ( 4.6, 1.6) {$C(\mathbf{w}_2)$};
\node at (-4.2,-4.6) {$C(\mathbf{w}_3)$};

\coordinate (yS)  at (-3.2,1.8);
\coordinate (ySi) at (-1.95,2.55);

\fill (yS) circle (2.5pt);
\fill (ySi) circle (2.5pt);
\draw[->, line width=1.5pt, gray!70!black] (yS) -- (ySi);

\node[ylab, above left=1pt]  at (yS)  {$\hat{y}_{S}$};
\node[ylab, above left=1pt]  at (ySi) {$\hat{y}_{S\cup\{i\}}$};

\node[text=gray!70!black, align=left] at (-3.4,1.2)
    {\textbf{No decision change}};

\coordinate (ySp)  at (3.05,-2.85);   
\coordinate (ySpi) at (1.8,-3.6);   

\fill (ySp) circle (2.5pt);
\fill (ySpi) circle (2.5pt);
\draw[->, line width=1.8pt, myblue] (ySp) -- (ySpi);

\node[ylab, right=1pt] at (ySp)  {$\hat{y}_{S'}$};
\node[ylab, left=2pt]      at (ySpi) {$\hat{y}_{S'\cup\{i\}}$};

\node[text=myblue, align=center] at (3.2,-2.35)
    {\textbf{Decision change}};

\end{tikzpicture}
    }
    \caption{Decision regions induced by the downstream optimization problem. The axes $\hat{y}_1, \hat{y}_2$ are the coordinates of the model prediction. In the center, the feasible set is defined by vertices $w_1, w_2, w_3$ with each colored region $C(w_k)$ being the normal cone of predictions that induce the corresponding decision $w_k$. While adding one feature to the coalition $S$ produces a change in the predicted vector $\hat{y}_S$, that change does not cross any decision boundaries, whereas for another coalition $S'$ it does. The decision value game isolates only the latter, namely those features that contribute to changes that shift between decision regions, and therefore impact the downstream policy.} 
    \label{fig:decision-vs-prediction}
\end{figure}
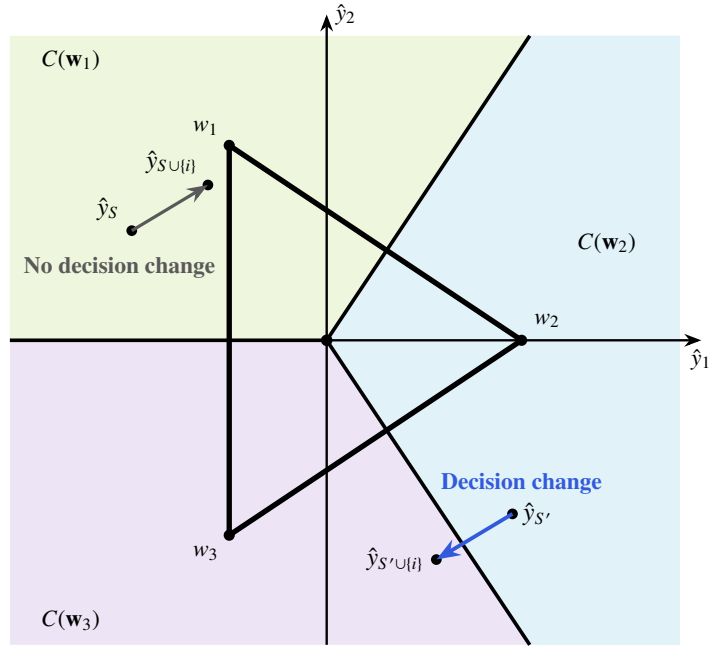

For an information player \(i\in\mathcal N\), the InfoDVA attribution is the Shapley value of the game in \cref{eq:infodva-game}:
\begin{equation}
\label{eq:infodva-shapley}
    \phi^{\mathrm{Info}}_{i,m}
    =
    \sum_{S\subseteq N\setminus\{i\}}
    \frac{|S|!(p-|S|-1)!}{p!}
    \left[
        v^{\mathrm{Info}}_{m}(S\cup\{i\})
        -
        v^{\mathrm{Info}}_{m}(S)
    \right].
\end{equation}
Because \cref{eq:infodva-shapley} applies the standard Shapley value to the decision-value game in \cref{eq:infodva-game}, InfoDVA inherits the usual Shapley properties for that fixed game. In particular, each feature is credited with its average marginal contribution to downstream decision value over all possible feature orderings, and the resulting attributions exactly sum to the full information decision value gain relative to the baseline. \cref{fig:decision-vs-prediction} gives a geometric interpretation of the marginal term in \cref{eq:infodva-shapley}. A feature can have a large predictive but zero decision-value effect when the induced prediction remains inside the same optimization decision region.
A positive value means that, on average over the order in which features are added, feature \(i\) improves decision value relative to the current coalition. A negative value means that the feature tends to induce decisions with lower value. In post-InfoDVA, this statement refers to realized outcomes, i.e., the actually realized gain/loss, while in the pre-InfoDVA it is taken with respect to the model's full prediction. Shapley efficiency ensures that the attributions sum to the total value \citep{shapley1953value}, so
\begin{equation}
\label{eq:infodva-efficiency}
    \sum_{i\in N}\phi^{\mathrm{Info}}_{i,m}
    =
    U\!\left(g_\rho(\hat y_N),\rho;\tilde y_m\right)
    -
    U\!\left(g_\rho(\hat y_{\emptyset}),\rho;\tilde y_m\right).
\end{equation}
Thus InfoDVA exactly decomposes the decision-value gain or loss from using the full information prediction instead of the baseline prediction.

\subsection{DesignDVA}

DesignDVA attributes decision value to design or optimization parameters, the configurable elements of the optimization model $g$. Let \( D=\{1,\ldots,q\}\) denote a set of design players. We consider a reference design \(\rho^0\) and a target design \(\rho^1\). For example, \(\rho^0\) might represent the current system, and \(\rho^1\) might represent a proposed alternative (larger battery, more vehicles, alternative routing). For a coalition \(T\subseteq D\), define the hybrid design,
\[
\rho_j(T) =
\begin{cases}
    \rho_j^1, & j \in T \\
    \rho_j^0, & j \notin T,
\end{cases}
\]
where the design coordinates in \(\rho\) are set to their target values and the remaining coordinates are kept at their reference values. If a design coordinate has more than two levels, the player represents the move from a specified reference level to a specified target level. Multi-level sensitivity analyses can be conducted by repeating the construction across reference--target pairs.

Holding the full prediction fixed, the DesignDVA game is
\begin{equation}
\label{eq:designdva-game}
    v^{\mathrm{Design}}_{m}(T)
    =
    U\!\left(g_{\rho(T)}(\hat y_N),\rho(T);\tilde y_m\right)
    -
    U\!\left(g_{\rho^0}(\hat y_N),\rho^0;\tilde y_m\right),
    \qquad
    T\subseteq D .
\end{equation}
The Shapley value of this game allocates the value of moving from the reference design to the target design across design coordinates. DesignDVA is useful for comparing operational configurations, but in this paper it primarily serves as a component of the joint information--design attribution developed next.

\subsection{JointDVA}

JointDVA attributes decision value jointly to information and design players. The player set is
$
    \mathcal P = N \cup D.
$
For any coalition \(A\subseteq\mathcal P\), evaluate the joint value function at $S=A\cap N$, $T=A\cap D.$ The coalition \(A\) therefore induces both a coalition prediction \(\hat y_S\) and a hybrid design \(\rho(T)\). The latter is a hybrid of the reference design \(\rho^0\) and \(\rho^1\), where coordinates corresponding to design players in $T$ are set to their target values, while the remaining stay at their reference values. For evaluation mode \(m\), define the JointDVA game
\begin{equation}
\label{eq:jointdva-game}
    v^{\mathrm{Joint}}_{m}(S,T)
    =
    U\!\left(
        g_{\rho(T)}(\hat y_{S}),
        \rho(T);
        \tilde y_m
    \right)
    -
    U\!\left(
        g_{\rho^0}(\hat y_{\emptyset}),
        \rho^0;
        \tilde y_m
    \right).
\end{equation}
The JointDVA attribution for player \(a\in\mathcal P\) is
\begin{equation}
\label{eq:jointdva-shapley}
    \phi^{\mathrm{Joint}}_{a,m}
    =
    \sum_{A\subseteq \mathcal P\setminus\{a\}}
    \frac{|A|!(|\mathcal P|-|A|-1)!}{|\mathcal P|!}
    \left[
        v^{\mathrm{Joint}}_{m}(A\cup\{a\})
        -
        v^{\mathrm{Joint}}_{m}(A)
    \right].
\end{equation}
JointDVA decomposes the full decision-value gain when moving from the baseline information--design pair \((\hat y_{\emptyset},\rho^0)\) to the target information--design pair \((\hat y_N,\rho^1)\):
\begin{equation}
\label{eq:jointdva-efficiency}
    \sum_{a\in\mathcal P}\phi^{\mathrm{Joint}}_{a,m}
    =
    U\!\left(g_{\rho^1}(\hat y_N),\rho^1;\tilde y_m\right)
    -
    U\!\left(g_{\rho^0}(\hat y_{\emptyset}),\rho^0;\tilde y_m\right).
\end{equation}
While the construction is simple, it is useful because it places information and design on a common operational scale. For example, the same decomposition can compare the decision value of a weather feature with the decision value of increasing battery capacity, or compare a demand-lag feature with the decision value of adding an EMS staging location. Further, it enables analyzing the interaction effect. 

\subsection{Decision-value interactions (DVI)}

Individual DVA scores allocate value to single players. In many PtO systems, however, the operational value of a player depends on the presence of another player. A feature may be valuable only under a sufficiently flexible design, or a design parameter may become valuable only when the predictive model has access to specific information. For our purposes, we refer to these as \emph{Decision-Value Interactions} (DVI).

For two players \(a,b\in\mathcal P\), coalition-specific pairwise interaction in coalition \(A\subseteq\mathcal P\setminus\{a,b\}\) is
\begin{equation}
\label{eq:second-order-difference}
    \Delta^{(2)}_{ab}v(A)
    = v(A\cup\{a,b\}) - v(A\cup\{a\})
    - v(A\cup\{b\}) + v(A).
\end{equation}
A positive value means that the incremental value of adding \(a\) and \(b\) together to \(A\) exceeds the sum of their separate incremental values relative to \(A\). A negative value means that their combined incremental value is less than that sum, indicating substitution or redundancy. To aggregate such effects across coalitions, we apply Faith-Shap interaction attribution to the relevant DVA game \citep{tsai2023faith}. Faith-Shap provides a way to turn coalition interaction terms in \cref{eq:second-order-difference} into a single interaction score. In the pairwise case, it compares the value of using two players together with the value of using each one separately, across possible coalitions, and averages these comparisons using the Shapley framework. Applied to a DVA game, the resulting score measures the extra decision value created by the pair itself, beyond the value attributed to the two players individually. For an interaction group \(G\subseteq\mathcal P\), \(|G|\ge 2\), define
\begin{equation}
\label{eq:dvi-definition}
    \Psi^{m} (G)
    =
    \mathrm{FaithShap}_{G}\!\left(v_m\right), \qquad G\subseteq P, |G| \geq 2,
\end{equation}
where \(v_m\) may be the InfoDVA, DesignDVA, or JointDVA characteristic function, depending on the question. In the case studies, we focus on pairwise interactions. We refer to \(\Psi^m(G)\) as DVI score. When \(G\) contains both an information player and a design player, we refer to it as \emph{cross-DVI} score.

Cross-DVI is the most relevant interaction object in this paper. It identifies when the usefulness of information is conditional on an operational design. For example, a solar-irradiance feature may interact positively with battery capacity if the information becomes valuable only when the battery can store enough energy to exploit midday price depressions. In an EMS coverage problem, a historical demand feature may interact positively with the staging location budget. When only a few ambulances can be staged, the optimizer may select the same high demand zones regardless of the input data. As the budget increases, however, the system has more deployment flexibility, so local demand information can help decide where the additional staging locations should be placed.

It is worth noting that DVI is not tied to any specific interaction attribution method. While we present it based on Faith-SHAP, since it appears to improve over other existing methods, such as the Shapley Interaction Index, the Banzhaf Interaction Index, Shapley-Owen, or Shapley-Taylor \citep{grabisch1999axiomatic,sundararajan2020shapley,owen1972multilinear,lundberg2020local}, any other Shapley-based interaction indices can be substituted without changing the DVA value function, only the allocation method of the interaction terms.
\subsection{Computation}
\label{subsec:compute-complexity}
As defined, DVA requires evaluating the characteristic function for all coalitions of the relevant player set. For player set $\mathcal P_v$ with \(|\mathcal P_v|\) players, this requires \(2^{|\mathcal P_v|}\) coalition evaluations. Each evaluation consists of constructing a coalition prediction, solving the downstream optimization problem under the corresponding design, and computing the decision value under the chosen evaluation mode. In the case studies, we choose feature groups and design groups so that exact enumeration is feasible.

At the same time, for the traditional Shapley method, there exists an active research into designing more computationally efficient approximations. By design, DVA can directly take advantage of many of such implementations \citep{lundberg2017unified,strumbelj2010efficient,mitchell2022sampling}. For example, permutation sampling estimates Shapley values by sampling orderings and evaluating marginal changes along each order. Kernel-style estimators sample coalitions and fit a weighted additive surrogate to the characteristic function. These approximations are directly applicable to DVA and are, in fact, quite effective. We present some of the results from our experimentation in \cref{subsec:shap-approximation-methods} and in Appendix B.

\section{Empirical Evaluation}
\label{sec:experiments}

\subsection{Overview and Setup}
\label{subsec:setup}
We use two diverse PtO demonstration cases, energy arbitrage presented in \cref{eq:caiso-arbitrage} and EMS maximum coverage in \cref{eq:ems-model}. Energy arbitrage is a classic problem in operations research \citep{jiang2015optimal,lohndorf2023value}. The specific case study of the day-ahead California market is especially popular due to the abundance of data and the potential impact a successful policy can have on overall profit \citep{byrne2018opportunities,krishnamurthy2017energy}. Emergency medical service (EMS) systems are also a natural setting for PtO decision support. Ambulance deployment and relocation decisions are important because emergency response resources are limited, demand is spatially and temporally heterogeneous, and small changes in staging decisions can affect whether calls are reachable within a target response radius. Prior work has studied ambulance dispatching, relocation, and coverage optimization in both real-time and strategic settings \citep{nasrollahzadeh2018real,boutilier2020ambulance,hashtarkhani2023place,frichi2025ambulance}. 

\begin{table}[ht]
\centering
\scriptsize
\caption{Overview of the PtO pipelines for the two empirical case studies.}
\label{tab:settings}
\begin{tabularx}{\textwidth}{@{}lXXXXX@{}}
\toprule
\textbf{Setting} & \textbf{Prediction $\hat y$} & \textbf{Decision} & \textbf{Features} & \textbf{Design parameters} & \textbf{Objective } \\
\midrule
CAISO arbitrage & 24-hour price & battery dispatch & weather \& calendar & battery capacity, battery efficiency & profit (\$/MWh) \\
EMS coverage & ZIP-hour demand & staging areas & demand, weather, time groups & budget and radius & covered demand (\%) \\
\bottomrule
\end{tabularx}
\end{table}

\cref{tab:settings} provides a high-level overview of the input features, design parameters, model outputs, and objectives of the two case studies. For both cases, we use a set of extreme gradient boosting tree models (XGBoost) as the predictors \citep{chen2016xgboost}. Tree-based models have been shown to be state-of-the-art for tabular datasets, often outperforming their deep learning counterparts \citep{grinsztajn2022tree,mcelfresh2023neural}. To assess whether the findings are sensitive to hyperparameter choices, we evaluate on 25 models, chosen through a \(L_{25}(5^6)\) orthogonal array experimental design with six factors across five levels \citep{montgomery2017design}. Implementation and dataset details are provided in Appendix A in the online supplement.

\paragraph{Energy Arbitrage Model} 
For each day, let the predictive model output a next-day hourly price vector
$\hat{p}_1, \dots, \hat{p}_{24}$. The energy arbitrage experiment uses hourly day ahead locational marginal prices from the CAISO OASIS system \citep{caiso_oasis_lmp_2026} for the SP15 trading hub with daily weather data from the Open-Meteo Historical Weather API \citep{zippenfenig_2024_14582479} using LAX as a proxy weather location for the SP15 region. Each observation corresponds to one operating day, and the predictive model maps weather and calendar features to a 24-dimensional vector of next-day prices, which is then passed to the battery dispatch model in \cref{eq:caiso-arbitrage}. We solve the following problem
\begin{subequations}
\label{eq:caiso-arbitrage}
\begin{alignat}{2}
\max \;\; & \sum_{t=1}^{24} \hat{p}_t\,(d_t - c_t)
    - \lambda \sum_{t=1}^{24} (c_t + d_t) & & \\
\text{s.t.} \quad
    & s_{t+1} = s_t + \eta_c\, c_t - \frac{1}{\eta_d}\, d_t
    &\quad& \forall\, t \\
    & 0 \le s_t \le E
    && \forall\, t \\
    & 0 \le c_t \le P^{\max}\, z_t
    && \forall\, t \\
    & 0 \le d_t \le P^{\max}(1 - z_t)
    && \forall\, t \\
    & z_t \in \{0,1\}
    && \forall\, t,
\end{alignat}
\end{subequations}
where $c_t$ is the charge power at hour $t$, $d_t$ is the discharge power at hour $t$, $s_t$ is the state of charge at hour $t$, $z_t$ is a binary variable selecting charge ($z_t=1$) or discharge ($z_t=0$) mode, $\eta_c, \eta_d$ are the charging and discharging efficiencies, $\lambda$ is a regularization parameter penalizing total throughput, $E$ is the energy capacity of the storage system, and $P^{\max}$ is the power limit (maximum charge/discharge rate). For this case study, we use $ \lambda = 5, \eta_c = \eta_d = 0.95, P^{\max} = 1, s_1=s_{25}=2, E=4$ unless stated otherwise.

\paragraph{EMS Maximum Coverage} For each explained hour \(t\), let the predictive model output a nonnegative
next-hour EMS demand score vector
\(\hat q_{ti}, \ i\in \mathcal I_t,\) and let $\hat{Q}_t = \sum_{i\in \mathcal I_t} \hat q_{ti}$ denote the total citywide EMS demand at hour $t$. The EMS coverage experiment uses New York City EMS incident dispatch records from NYC Open Data \citep{ems2016}, aggregated to hourly incident counts by Manhattan ZIP code zones. The prediction task is then to forecast hourly EMS demand across these 45 zones using time, weather, lagged demand, neighboring-zone demand, and historical zone-hour baseline features. Then we solve the budgeted maximum coverage problem
\begin{subequations}
\label{eq:ems-model}
\begin{alignat}{2}
\max \;\; & \frac{\sum_{i\in \mathcal{I}_t} \hat{q}_{ti}\, y_i}{\hat{Q}_t}
    & & \\
\text{s.t.} \quad
    & y_i \le \sum_{j\in \mathcal{J}_t} a_{ij}\, x_j
    &\quad& \forall\, i \in \mathcal{I}_t \\
    & \sum_{j\in \mathcal{J}_t} x_j \le p
    && \\
    & x_j \in \{0,1\}
    && \forall\, j \in \mathcal{J}_t \\
    & y_i \in \{0,1\}
    && \forall\, i \in \mathcal{I}_t,
\end{alignat}
\end{subequations}
where \(x_j\) is a binary variable indicating whether candidate staging location \(j\) is selected, \(y_i\) is a binary variable indicating whether demand zone \(i\) is covered, \(\hat q_{ti}\) is the predicted next-hour EMS demand score for zone \(i\) at hour \(t\), \(\mathcal I_t\) is the set of demand zones considered at hour \(t\), \(\mathcal J_t\) is the set of candidate EMS staging locations considered at hour \(t\), \(a_{ij}\) is a binary coverage parameter denoting if staging location \(j\) can cover demand zone \(i\), and \(p\) is the maximum number of staging locations that may be selected.
The coverage parameter is defined by
\[
a_{ij}
=
\begin{cases}
1, & \text{if the distance from staging location } j
     \text{ to demand zone } i \text{ is at most } \tau \text{ away}, \\
0, & \text{otherwise,}
\end{cases}
\]
where \(\tau\) is the response-time threshold radius in km.

\subsection{Metrics \& Baseline Performance}
\label{subsec:metrics}
The case studies compare DVA with predictive feature-attribution baselines. Predictive SHAP is computed on the predictive model output and then aggregated to a scalar score when the model output is vector-valued. Decision-aware baselines include leave-one-feature-out (LOFO) decision ablation, downstream permutation feature importance, and greedy decision insertion. LOFO scores each feature \(i\) by the decision value loss from removing it from the full feature set and ranks features by this resulting loss \citep{covert2021explaining}. Downstream permutation feature importance shuffles each feature in the evaluation set, recomputes the decision value loss, and ranks features by the average drop in decision value \citep{fisher2019all}. Greedy decision insertion constructs a feature ordering sequentially, starting from the empty set and repeatedly adding features with the largest average improvement in decision value over the evaluation set \citep{petsiuk2018rise}.

We use two metrics. First, decision infidelity measures whether an attribution vector approximates the decision-value loss associated with removing or perturbing information. Lower values indicate a more faithful magnitude explanation \citep{ye2019infidelity}. Second, decision insertion AUC (area under the curve) evaluates the ranking induced by an attribution method. Starting from the null coalition, players are added in attribution order, and we record how quickly decision value is recovered. Higher insertion AUC indicates that the attribution ranking identifies better rankings of decision-relevant players \citep{wang2024benchmarking,rong2022consistent}.

First, to establish predictive and prescriptive model quality, in \cref{tab:aggregate-objective-performance-ci} we show the aggregate performance of the XGBoost models in terms of the predictive power and realized downstream decisions (compared to the perfect information solution, referred to as oracle). Next, in \cref{tab:combined_attribution_metrics}, we compare the quality of DVA attributions, specifically post-InfoDVA, to traditional SHAP and the decision-aware methods introduced in \cref{subsec:metrics} in both ranking quality and magnitude. Observe that DVA outperforms or is competitive with all baselines in ranking quality on average, represented by the decision insertion AUC scores. Through decision infidelity we also show that DVA provides reliable approximations for downstream objectives across feature coalitions. Since prediction SHAP is based on model outputs, it cannot reliably approximate the downstream objective, whereas DVA is a much more faithful approximator. In fact, in terms of decision infidelity, Prediction SHAP is entirely uninformative, performing orders of magnitude worse, while DVA scores well. Note that since the decision-aware baselines (LOFO, permutation, and greedy insertion) only offer rankings of features, we cannot use them to compute the decision infidelity metric.

\begin{table}[ht]
\centering
\small
\caption{Aggregate predictive and realized objective performance across 25 models with 95\% bootstrap confidence intervals for the case studies. Here the oracle represents the objective value with perfect information.}
\label{tab:aggregate-objective-performance-ci}
\begin{threeparttable}
\begin{tabular}{lrr}
\toprule
Case study & RMSE & Objective gap vs Oracle\tnote{\dag} \\
\midrule
CAISO & 24.066 [23.431, 24.695] & 10.858 [9.983, 11.805] \\
EMS ($\tau=1,\ p=3$) & 1.024 [1.019, 1.031] & 12.134\% [11.876\%, 12.392\%] \\
EMS ($\tau=1,\ p=5$) & 1.024 [1.019, 1.031] & 15.445\% [14.957\%, 15.965\%] \\
EMS ($\tau=1,\ p=8$) & 1.024 [1.019, 1.031] & 17.054\% [16.686\%, 17.469\%] \\
EMS ($\tau=2,\ p=5$) & 1.024 [1.019, 1.031] & 5.437\% [5.369\%, 5.507\%] \\
\bottomrule
\end{tabular}
\begin{tablenotes}
\small
\item[\dag] CAISO: \$/MWh; EMS: percentage-point coverage gap.
\end{tablenotes}
\end{threeparttable}
\end{table}

\begin{table}[ht]
\centering
\scriptsize
\caption{Attribution metric comparison across methods for CAISO and EMS with 95\% paired bootstrap confidence intervals on 30 evaluation samples. Values are mean [CI].}
\label{tab:combined_attribution_metrics}
\begin{tabular}{llcc}
\toprule
\textbf{Metric} & \textbf{Method} & \textbf{CAISO} & \textbf{EMS} \\
\midrule
\multirow{5}{*}{Decision insertion AUC $\uparrow$}
  & \textbf{Post-InfoDVA} & \textbf{0.631 [0.508, 0.744]} & \textbf{0.801 [0.750, 0.845]} \\
  & Prediction SHAP & 0.475 [0.375, 0.578] & 0.656 [0.557, 0.749] \\
  & Leave-one-feature-out & 0.599 [0.490, 0.699] & 0.675 [0.567, 0.765] \\
  & Greedy decision insertion & 0.544 [0.429, 0.658] & 0.699 [0.602, 0.780] \\
  & Permutation feature importance & 0.292 [0.179, 0.404] & 0.481 [0.363, 0.586] \\
\midrule
\multirow{2}{*}{Decision infidelity $\downarrow$}
  & \textbf{Post-InfoDVA} & \textbf{36.04 [27.22, 48.19]} & \textbf{0.0003 [0.0002, 0.0004]} \\
  & Prediction SHAP & 34,195.79 [24,750.40, 46,003.04] & 38.22 [25.43, 51.89] \\
\bottomrule
\end{tabular}
\end{table}

\subsection{Design parameters drive information value}

We next present results of an experiment aimed at answering whether predictive information remains operationally useful under different problem parameters. Here, we focus on the EMS design problem. A conventional sensitivity analysis can compare realized coverage across different optimization parameters, but it cannot determine whether input information is being used for downstream decision value, nor which feature--design combinations create that value. We evaluate our EMS maximum coverage dataset on 9 regimes in a 3x3 design, for coverage radii $\tau \in \{ 1,2,3\}$, and a facility budget $p \in \{3,5,8\}$. These regimes allow us to distinguish three types of downstream behavior. We call a regime \emph{decision-active} when changes in the input features frequently change the selected EMS staging policy and its realized coverage value. A regime is \emph{decision-stable} when the policy is insensitive to input features, even though the resulting coverage remains below the best attainable level. A regime is \emph{saturated} when the policy is insensitive because the design constraints are relaxed enough that the system already achieves close to complete coverage. Thus, the experiment identifies whether predictive information is useful, ignored by the PtO pipeline, or rendered unnecessary by the design.

\begin{figure}[ht]
    \centering
    \includegraphics[width=\linewidth]{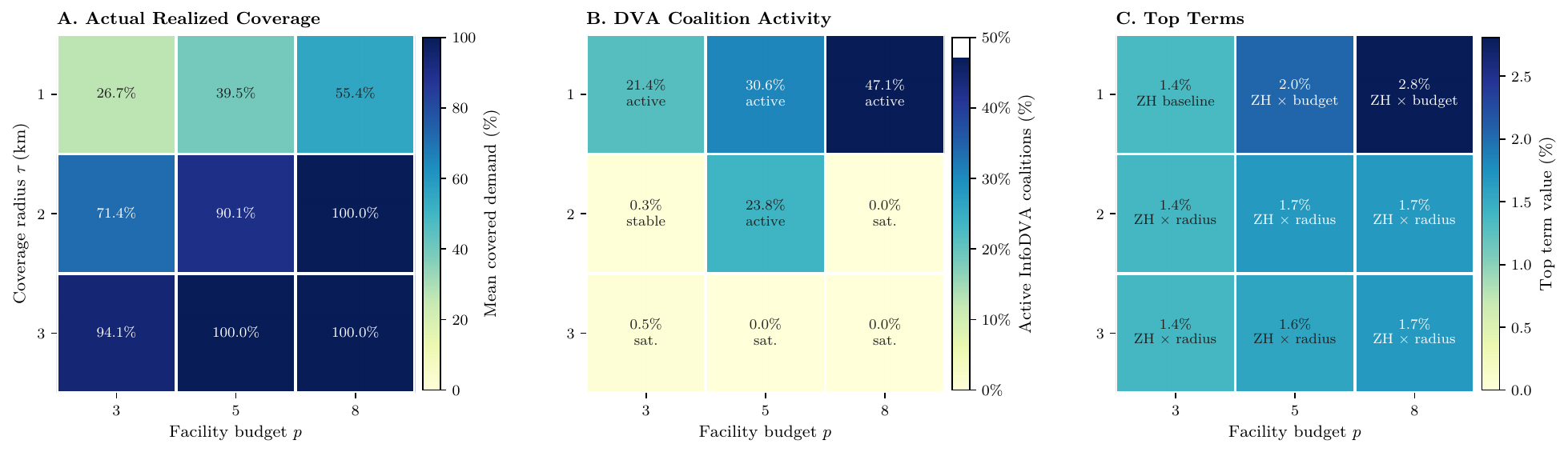}
    \caption{EMS decision regime heatmaps. Panel A reports realized covered demand across nine budget-radius regimes. Panel B reports the fraction of InfoDVA coalition evaluations with nonzero realized decision-value change. Panel C reports the dominant information–design interaction in each regime using post-JointDVI, with values expressed in percentage points of realized coverage.}
    \label{fig:ems-decision-regimes}
\end{figure}

In \cref{fig:ems-decision-regimes}A, we present the conventional performance view of realized coverage across budgets and radii, which can be interpreted as traditional sensitivity analysis. Panel B presents the DVA perspective, which measures how often feature coalitions in the InfoDVA game produce nonzero decision-value changes. Several regimes have near-zero decision-active coalition rates, but for different reasons. When $\tau = 3$ or $\tau = 2, p = 8$, the near-zero activity coincides with near-complete coverage. By contrast, when $\tau = 2$ and $p = 3$, the decision-value change is also near zero, but the realized coverage is below saturation. This is referred to as decision-stable regime, where the design maps many different feature coalitions to the same realized coverage outcome, even though substantial demand remains uncovered. This coincides with actual decision changes across coalitions. In regime $\tau =2, p=3$, both the policy ($0.44\%$) and the value ($0.3\%$) remain near-constant, meaning the PtO pipeline has identified three specific staging areas as optimal choices regardless of any input features, i.e., these three areas are optimal for all likely demand realizations. 

\cref{fig:ems-decision-regimes}C explains where the decision-active behavior comes from. The dominant post-JointDVI terms show that the key feature is the zone-hour baseline (ZH): a rolling average of incidents for each zone-hour pair. ZH does not have a fixed value across designs. Instead, its value is mediated by budget and coverage radius. In low radius regimes additional budget makes spatial information more actionable. In larger radius regimes radius interactions dominate, and the overall impact of input features decreases (JointDVI $5.8\% \rightarrow 2.4\%$) as saturation increases. A more detailed view can be found in \cref{fig:ems-joint-dvi-detailed} in the Appendix. DesignDVA and DVI corroborate these findings, as shown in \cref{tab:ems-designdva-with-interactions}. The change from $\tau = 1 \rightarrow 2$ results in the largest coverage gain of $44.72\%$, with diminishing gains at $\tau = 3$. In the case of $\tau = 2, p = 5$ the decision gain interaction is complementary, with an additional $5.94\%$ coverage with both parameters improved, but negative otherwise, due to saturation.

\begin{table}[h]
\centering
\small
\sisetup{table-format=2.2, table-number-alignment=center}
\caption{EMS DesignDVA and DVI decomposition of actual realized coverage. Values are signed percentage points relative to the baseline design $\tau = 1, p = 3$ with a baseline coverage of $\phi_\emptyset = 26.72\%$, averaged across 25 models. $\phi_\tau$ represents the percentage coverage attributed to radius $\tau$, $\phi_p$ the attribution to number of staging areas $p$, and $I_{\tau , p}$ the DVI interaction attribution.}
\label{tab:ems-designdva-with-interactions}
\begin{tabular}{@{}ll SSS@{}}
\toprule
& & {$p=3$} & {$p=5$} & {$p=8$} \\
\midrule
\multirow{3}{*}{$\tau=1$}
  & $\phi_\tau$  & 0.00 & 0.00  & 0.00  \\
  & $\phi_p$     & 0.00 & 12.74 & 28.69 \\
  & $I_{\tau,p}$ & 0.00 & 0.00  & 0.00  \\
\midrule
\multirow{3}{*}{$\tau=2$}
  & $\phi_\tau$  & 44.72 & 47.69 & 44.65 \\
  & $\phi_p$     & 0.00  & 15.71 & 28.62 \\
  & $I_{\tau,p}$ & 0.00  & {+5.94} & {-0.14} \\
\midrule
\multirow{3}{*}{$\tau=3$}
  & $\phi_\tau$  & 67.34 & 63.94 & 55.96 \\
  & $\phi_p$     & 0.00  & 9.34  & 17.31 \\
  & $I_{\tau,p}$ & 0.00  & {-6.80} & {-22.75} \\
\bottomrule
\end{tabular}
\end{table}

Overall, we can observe that, in this case, DVA can allow the decision maker to identify which regimes are worth explaining and which design parameter combinations result in input insensitivity. While regular regret comparison can identify where the model does not add decision value, DVA allows the user to connect that loss to specific input features and design parameters, and show, through pre/post-DVA comparison, which harmful policy changes are improvements under the model's prediction. 
In our case, the regimes with $\tau = 1$ and $\tau = 2, p = 5$ have downstream policies that are responsive to different inputs.
\subsection{Identifying disruptive features}
\label{subsec:results-gdva}

\begin{figure}[ht]
    \centering
    \includegraphics[width=\linewidth]{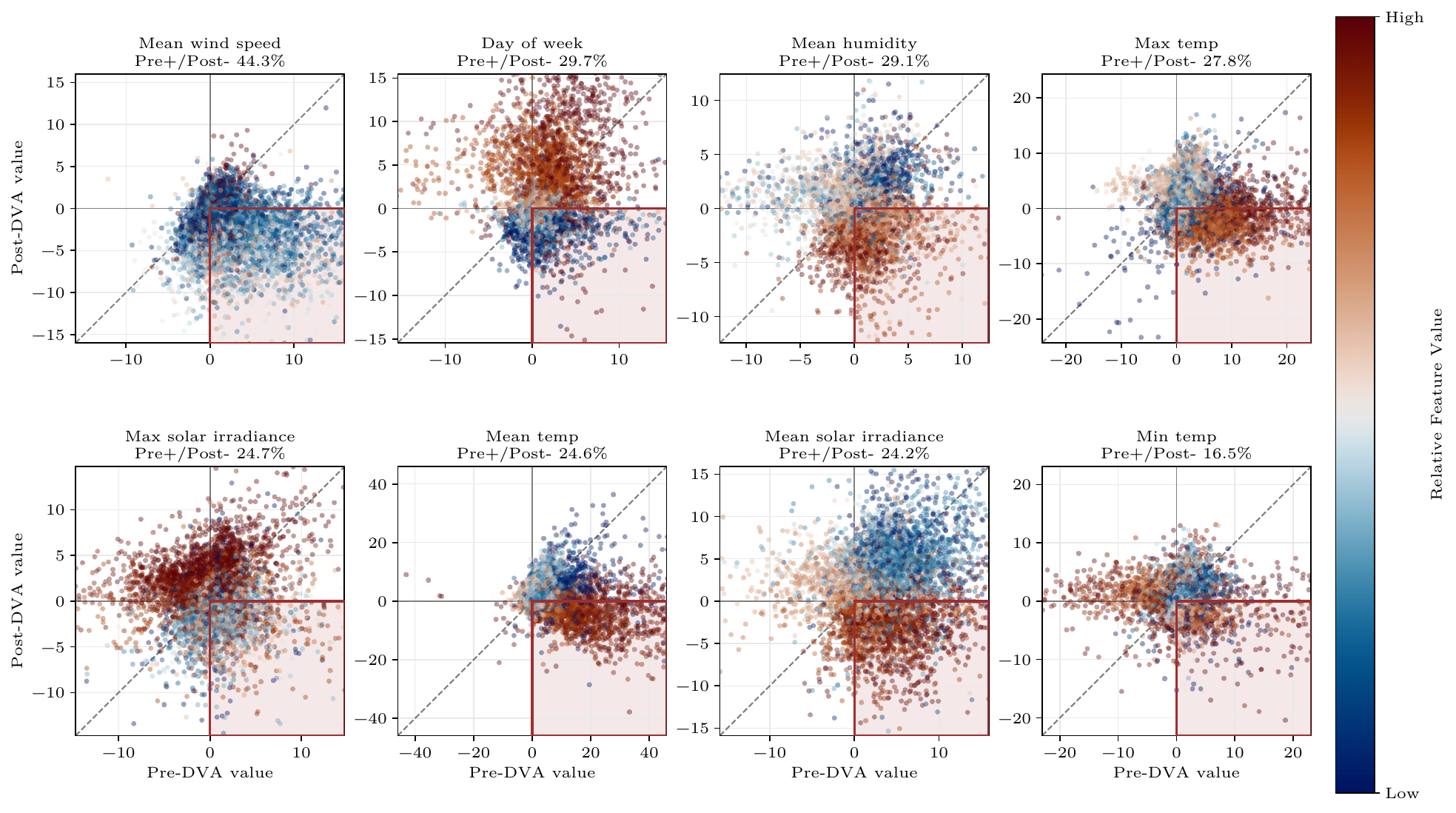}
    \caption{Pre-InfoDVA / Post-InfoDVA value comparison across features in the validation set. The highlighted quadrant is the set of points where the feature provides positive decision value according to the model predictions (pre-DVA) but negative realized decision value (post-DVA). 
    }
    \label{fig:scatter-pre-post}
\end{figure}

DVA can help diagnose failures in PtO pipeline that classic interpretation frameworks may miss. We begin with a comparison of pre- and post-InfoDVA in the CAISO case study, shown in \cref{tab:caiso-pre-post-infodva}. 
\begin{table}[ht]
\centering
\scriptsize
\caption{Comparison of pre-InfoDVA and post-InfoDVA. We report the mean absolute values for both, the mean absolute value of their difference as disagreement, and the percentage of samples where pre-InfoDVA had a positive value but post-InfoDVA had a negative value. Results are reported with 95\% paired bootstrap confidence intervals.}
\label{tab:caiso-pre-post-infodva}
\begin{tabular}{lcccc}
\toprule
Feature & Mean abs Pre-DVA & Mean abs Post-DVA & Disagreement & Pre+/Post$-$ rate \\
\midrule
Mean wind speed & 2.64 [2.55, 2.73] & 2.31 [2.26, 2.36] & 4.03 [3.91, 4.15] & 44.3\% [43.3, 45.4] \\
Day of week & 1.77 [1.70, 1.83] & 2.40 [2.33, 2.47] & 2.75 [2.66, 2.83] & 29.7\% [28.8, 30.7] \\
Mean humidity & 1.91 [1.85, 1.98] & 1.74 [1.70, 1.78] & 2.58 [2.51, 2.66] & 29.1\% [28.2, 30.0] \\
Max temp & 4.19 [4.06, 4.34] & 2.92 [2.86, 2.97] & 4.88 [4.74, 5.03] & 27.8\% [26.9, 28.7] \\
Max solar irradiance & 2.36 [2.28, 2.45] & 2.04 [1.99, 2.08] & 3.10 [3.01, 3.20] & 24.7\% [23.8, 25.6] \\
Mean temp & 7.12 [6.87, 7.37] & 4.32 [4.24, 4.40] & 7.66 [7.38, 7.94] & 24.6\% [23.8, 25.5] \\
Mean solar irradiance & 3.77 [3.69, 3.85] & 3.34 [3.28, 3.40] & 3.94 [3.85, 4.03] & 24.2\% [23.3, 25.1] \\
Min temp & 3.20 [3.01, 3.41] & 2.22 [2.18, 2.27] & 3.83 [3.61, 4.06] & 16.5\% [15.8, 17.3] \\
\bottomrule
\end{tabular}
\end{table}
Pre-InfoDVA represents the feature importance attribution under the model's prediction. Post-InfoDVA represents the feature importance attribution after observing the actual result. If pre-DVA believes a feature to be a positive source of information for downstream decision value, but post-DVA shows it is actually consistently detrimental, that can help diagnose a flaw in the predictive model, especially if that feature is influential (represented by a high absolute value). Here we see that \emph{mean wind speed} has a Pre+/Post- disagreement rate of 44.3\%, significantly higher than other features. This means that for 44.3\% of the explained samples, \emph{mean wind speed} was detrimental to decision value, even though pre-DVA identified it as positive and influential. Therefore, across the set of 25 models, the estimation error that \emph{mean wind speed} is associated with leads to dispatch decisions that have lower realized value. To visualize this disagreement, \cref{fig:scatter-pre-post} shows pre-InfoDVA against post-InfoDVA for each feature over the validation set. Each panel represents a two-dimensional decision value \textit{beeswarm} plot, where points to the right of zero are features that improve decision value under the model's own prediction, and points below zero are features that reduce the realized decision value. The highlighted lower-right quadrant therefore identifies cases where a feature appears useful under the model's prediction but induces decisions with negative decision value. We also present a traditional beeswarm in \cref{fig:caiso-beeswarm} in the Appendix.

Based on this finding, we devise an automated criterion for identifying disruptive features in models based on post-InfoDVA, and use it to guide a post hoc intervention aimed at improving downstream regret, which we call Guided DVA intervention (GDVA). We train each model on the 24-month training set, and compute InfoDVA values on the 364-day validation set (one day removed due to missing data). We then identify up to three features as potentially disruptive using the following criteria:
\begin{itemize}
    \item The feature must be detrimental to the model's realized decision gain. The mean InfoDVA value of the feature over the validation set must be negative.
    \item The feature must be influential to downstream decisions. The mean absolute InfoDVA value must be greater than 1\% of the maximum validation mean absolute InfoDVA value across features.
    \item The feature must be consistently disruptive to the full model. The 95\% bootstrap upper confidence interval of the feature's InfoDVA value must also be negative.
\end{itemize}

For each qualifying feature, we compute the mean actual daily regret improvement on the validation set when that feature is masked from the model. The feature with the largest improvement is then chosen to be masked on the test set of 101 days, and compared with the full model (FM). We compare against random feature masking, where a random feature that does not fulfill the GDVA criteria is masked instead. Finally, we also compare with SAGE, which uses Shapley values to attribute model loss (here MSE) to individual features, and use the same procedure to identify features that are disruptive to the predictive power of the model \citep{covert2020understanding}. The results are shown in \cref{tab:caiso-guided-neutralization}.

\begin{table}[t]
\small
\centering
\caption{Out-of-sample effect of the Guided DVA (GDVA) intervention.
For each of the 25 models, the feature to mask is selected using validation-period post-InfoDVA only. The selected feature is then replaced by its background expectation and evaluated once on the held-out test period. 
Positive $\Delta$ DV (decision value) and $\Delta$ RMSE indicate improvement over the comparison methods. Uncertainty intervals are 95\% bootstrap confidence intervals.}
\label{tab:caiso-guided-neutralization}
\begin{tabular}{lccc}
\toprule
\multicolumn{4}{l}{\textbf{Aggregate test-period intervention performance}} \\
\midrule
Comparison & Test Wins & Mean $\Delta$ DV 
    & Mean $\Delta$ RMSE \\
\midrule
GDVA vs. FM 
    & 21/23 & 1.63 [1.06, 2.24] & 1.98 [1.35, 2.65] \\
GDVA vs. random
    & 19/23 & 1.50 [0.85, 2.20] & 0.70 [-0.30, 1.75] \\
GDVA vs. SAGE
    & 18/23 & 1.10 [0.54, 1.72] & -1.67 [-2.09, -1.23] \\
\midrule
\multicolumn{4}{l}{\textbf{Features selected by the GDVA validation rule}} \\
\midrule
Selected feature & Test Wins & Mean $\Delta$ DV & Mean $\Delta$ RMSE \\
\midrule
Mean wind speed        & 20/22 & 1.68 [1.08, 2.30] & 2.03 [1.38, 2.73] \\
Max temperature        & 1/1  & 0.72 [0.72, 0.72] & 0.88 [0.88, 0.88] \\
\bottomrule
\end{tabular}
\end{table}

We see that through this process and across different model specifications, a feature was chosen 23/25 times, and \emph{mean wind speed} was chosen 22 times, while max temperature was chosen once. GDVA outperformed the full model 21/23 times, with a decision gain of 1.63 \$/MWh. GDVA also outperformed random masking 19/23 times, and outperformed SAGE 18/23 times. SAGE masking resulted in a more accurate model ($\Delta$ RMSE of $-1.67$) but that did not translate to better decision value. Overall, we conclude that through analysis on a holdout validation set, DVA was able to identify disruptive features across the entire PtO pipeline on a test set where other methods could not. 
\subsection{SHAP Approximation Methods}
\label{subsec:shap-approximation-methods}
Exact DVA is exponential in the number of players or features, and therefore, approximation methods are often required for large real-world datasets. Here, we adapt two popular approximation methods, Kernel SHAP and Permutation SHAP, discussed in \cref{subsec:compute-complexity}. We report the Kendall tau-b similarity of the global rankings produced and the agreement percent of the global top-1 feature with the exact DVA attribution \citep{kendall1938new,kendall1945treatment}. We also compute a normalized MAE (NMAE) of the DVA values from the exact values, to test how well the exact values are approximated by each method. We report the results in \cref{tab:ems-approx-decision-shap}. First, observe that the approximation methods substantially reduce the computational cost of DVA. Next, note that Permutation SHAP recovers the global ranking well even at small budgets (Kendall tau $\approx 0.93$ with 96\% top-1 agreement at 16 sampled orderings), and both estimators reach near-exact recovery (Kendall tau $\approx 0.97$) at 73–76\% of the exact runtime. Full results are reported in \cref{tab:ems-approx-decision-shap-full} in the Appendix.
\begin{table}[h]
\centering
\caption{Approximation quality for EMS InfoDVA in the decision-active design $(p=8,\tau=1)$, where exact InfoDVA enumerates all $2^8=256$ coalitions per explained hour. Time is the percentage of exact runtime, and Top-1 is \% agreement with the exact method on the most important feature per sample. Values are averaged over 50 runs per method per budget.}
\label{tab:ems-approx-decision-shap}
\small
\begin{tabular}{@{}llrrrr@{}}
\toprule
Estimator & B & Time (\% exact) & NMAE & Kendall $\tau$ & Top-1 \\
\midrule
Exact       & 256 & 100.0\% & ---   & ---   & ---   \\
\midrule
Kernel      & 16  & 7.0\%   & 1.441 & 0.749 & 76\%  \\
Kernel      & 192 & 75.8\%  & 0.222 & 0.966 & 100\% \\
\midrule
Permutation & 16  & 32.5\%  & 0.473 & 0.929 & 96\%  \\
Permutation & 64  & 73.2\%  & 0.238 & 0.967 & 100\% \\
\bottomrule
\end{tabular}
\end{table}

\section{Discussion \& Conclusions}
\label{sec:discussion}

We introduced Decision-Value Attribution, a Shapley-based framework for explaining PtO systems by decomposing downstream decision value rather than predictive output. By defining games over information sources, design parameters, and their interactions, DVA places features, operational choices, and solvers on a common objective scale. Across electricity arbitrage and EMS coverage, DVA identifies when predictive explanations are poor proxies for decision value, when information should be controlled or simplified, and when design choices determine whether information is useful. These results suggest that explanations for operational AI systems should be judged by whether they clarify and improve the decisions those systems produce, rather than based on the quality of the predictions. 

Our results suggest that explanations for PtO systems should take into account the entire PtO pipeline to be reliable. Predictive attribution methods identify variables that move the model output, but the optimizer can filter the value of those changes. DVA addresses this gap by attributing realized or model-implied decision value in the units of the downstream objective. This distinction is demonstratively important in both case studies, where features that are influential for prediction are not necessarily valuable for realized dispatch or coverage decisions, while simple baseline signals can dominate the value of the induced policy. In fact, our experiments show that pre-DVA is most valuable when complementing post-DVA. Since it represents the decision value based on the model prediction, large disagreement between pre- and post-DVA values can reveal opportunities for interventions or highlight flaws in the predictive model. 

The results also illustrate how DVA can be used in diagnosis. In the CAISO case, post-DVA identifies features that are influential but consistently harmful to realized decision value, leading to a targeted GDVA intervention that improves held-out decision performance. In the EMS case, the comparison between pre- and post-DVA reveals that the full model often changes staging decisions in ways the model regards as nearly equivalent but that are materially different (and often deleterious) under realized demand. This diagnosis motivates a simpler baseline-driven policy, which performs as well as or better than the full model out of sample. Thus, DVA can support both feature-control interventions and the discovery of simpler interpretable policies. 

A second implication is that DVA can reveal when predictive information is operationally useful. In the EMS study, the same predictive model and feature set have different decision value depending on the budget, coverage radius, and solution procedure. DVA identifies regimes in which the PtO pipeline is decision-active, where information can meaningfully change the selected policy. This distinction is useful because it tells the analyst whether poor performance should be addressed by improving the predictive model, changing the optimization design, or simplifying the policy. DVI extends this diagnostic by showing which information sources become valuable only under particular design choices, such as specific budget-radius combinations. 

The framework has important limitations. First, DVA inherits the dependence of Shapley-based explanations on the definition of missing information. In this paper we use empirical background marginalization, which is standard in SHAP attribution, but different coalition constructions, especially conditional constructions that preserve feature dependence, may produce different attributions. Therefore, DVA score interpretations depend on the distribution of the chosen background dataset. Second, the choice of players matters. Feature grouping and design-parameter grouping are often necessary for interpretability and computation, but different groupings answer different managerial questions and may redistribute value across components. Third, exact DVA scales exponentially in the number of players because each coalition requires a prediction, an optimization solve, and a value evaluation. Approximation methods can reduce this computational burden, but then the attributions inherit sampling error and may be less reliable for small or unstable effects. Finally, DVA is a diagnostic tool and does not imply causality. Post-DVA can identify features or design choices that were associated with higher or lower realized decision value for a PtO pipeline, but it does not by itself prove that a specific intervention will improve future performance. 

Several directions remain open for future research. First, although exact DVA is feasible for the grouped player sets studied here, larger PtO systems will require scalable approximation methods. Future work could develop solver-aware DVA estimators that utilize both quicker heuristics and exact solvers to provide reliable DVA value approximations. Another interesting direction is extending the framework to analyze different aspects of the PtO pipeline. Evaluating the impact of different heuristics on decision value, and their interactions with input information relative to an exact baseline can provide useful insights into the decision-making process of heuristics and approximate solvers. Another possibility is to shift the focus from diagnosis to intervention. The GDVA results show that post-DVA can identify features that are harmful for a fixed pipeline, but further work is needed to understand when DVA-guided interventions remain effective after retraining. More generally, DVA could be integrated into model-selection and decision-focused learning procedures, so that prediction models are trained to both improve downstream value but also to avoid misleading decisions.

\section{Appendices \& Supplementary Data}
\label{sec:app-data}
We provide detailed implementation details in Appendix A and additional figures, compute time comparisons, and detailed SHAP approximation results in Appendix B. Both Appendices are submitted for review as a separate supplementary file. We provide the code for reproducing our results, along with the processed datasets, in \href{https://github.com/kziliask/DVA}{\texttt{github.com/kziliask/DVA}}.

\section*{Acknowledgements}
This work was made possible in part by a grant of high performance computing resources and technical support from the Alabama Supercomputer Authority.

\bibliographystyle{elsarticle-harv}
\bibliography{bib}
\section*{Appendix A. Implementation Details}
\subsection*{XGBoost experimental setup}
\label{appsubsec:xgboost-robustness}
To assess whether the findings were sensitive to hyperparameter choices, we evaluated 25 model specifications. The specifications were generated using an \(L_{25}(5^6)\) orthogonal array design with six factors, each evaluated at five levels. Through this setup, each level of each hyperparameter appears five times, and every pairwise combination of levels across any two hyperparameters appears once. The considered factors are as follows.
\begin{itemize}
    \item \textbf{Number of trees} controls how many sequential decision trees are added to the ensemble, with more trees generally allowing more complex fitted patterns.
    \item \textbf{Maximum tree depth} controls how many splits each tree can make, with deeper trees capturing more detailed interactions but increasing the risk of overfitting.
    \item \textbf{Learning rate} controls how strongly each new tree updates the model, with smaller values making learning more gradual and often requiring more trees.
    \item \textbf{Row subsampling rate} controls the fraction of training observations used to fit each tree, which can reduce overfitting by adding randomness.
    \item \textbf{Column subsampling rate} controls the fraction of predictors considered when fitting each tree, which can reduce reliance on any single feature and improve generalization.
    \item \textbf{\(L_2\) regularization strength} penalizes large leaf weights, making the model more conservative and helping reduce overfitting.
\end{itemize}
The six varied hyperparameters chosen and their corresponding levels are shown in \cref{tab:xgboost-levels}. The first level in each hyperparameter represents the default XGBoost configuration. All other preprocessing, training, and evaluation choices were held fixed across the 25 specifications.
\begin{table}[ht]
\centering
\small
\caption{Hyperparameter levels used in the XGBoost robustness design}
\label{tab:xgboost-levels}
\begin{tabular}{ll}
\toprule
\textbf{Hyperparameter} & \textbf{Levels} \\
\midrule
Number of trees & 100, 50, 150, 250, 350 \\
Maximum tree depth & 3, 2, 4, 5, 6 \\
Learning rate & 0.05, 0.01, 0.03, 0.10, 0.15 \\
Row subsampling rate & 0.90, 0.60, 0.75, 0.95, 1.00 \\
Column subsampling rate & 0.90, 0.60, 0.75, 0.95, 1.00 \\
\(L_2\) regularization strength & 1.0, 0.3, 3.0, 10.0, 30.0 \\
\bottomrule
\end{tabular}
\end{table}
\subsection*{Feature Engineering and Preprocessing}
\label{appsubsec:feature-engineering}
\subsubsection*{CAISO Case Study}
The CAISO dataset contains hourly day-ahead market prices for the SP15 pricing region, sourced from CAISO Open Access Same-Time Information System (OASIS) \citep{caiso_oasis_lmp_2026}. SP15, or South of Path 15, is a CAISO aggregated trading hub associated with the region south of the Path 15 transmission corridor, a major north--south transmission interface in California, and is representative of wholesale energy prices in Southern California \citep{eia_daily_energy_prices}. The processed dataset spans 1,194 usable daily rows from January 26, 2023 through May 7, 2026. The calendar split is January 26, 2023 through January 25, 2025 for training, January 26, 2025 through January 25, 2026 for validation, and January 26, 2026 through May 7, 2026 for testing, yielding 729 training days, 364 validation days, and 101 test days after missing-price days are removed. Four rows are removed because pricing data is missing on March 12, 2023, March 10, 2024, March 9, 2025, and March 8, 2026.

The CAISO price data points are operating-hour LMP energy prices for SP15 in the day-ahead market. Following prior work showing that weather covariates are informative for energy-price prediction \citep{lamp2022large,mandl2022data}, the model covariates are daily minimum, mean, and maximum temperature, daily mean humidity, daily mean wind speed, day of the week, and daily mean and maximum solar irradiance. The weather data are daily summaries from the Open-Meteo Historical Weather API \citep{zippenfenig_2024_14582479}. The weather point is LAX, which we use as a proxy for the SP15 zone. The code uses one weather vector for the SP15 hub because the SP15 target is an aggregate trading-hub price and the Open-Meteo LAX record covers the CAISO sample dates. The day of the week feature is coded with Monday equal to 0.

The daily price target is a 24-dimensional vector, so the model output $\hat{p}_t \in \mathbb{R}^{24}$ represents the predicted day-ahead prices over the 24 operating hours of day $t$. The XGBoost model is trained with squared-error loss, histogram trees, under the experimental design shown in \cref{tab:xgboost-levels}. DVA values are computed on the 364-day validation set, and predictive and downstream decision metrics are reported on the 101-day test holdout set. The CAISO background set is the last 365 calendar days in the training frame. As per the Shapley method, for a coalition, the implementation copies the background feature matrix, replaces included feature columns by the explained day's values, predicts all background rows, and averages the resulting 24-hour price vectors.

For our default setup, we use 4-hour capacity batteries, which align with large scale CAISO operations \citep{caiso2025battery}. According to \citet{cole2025battery}, typical battery efficiency values ranges from 80-90\% round trip, which aligns with our choice of $\eta_d = \eta_c =95\%$, $0.95^2 \approx 0.9$. Finally, a throughput penalty is used in modeling as a battery degradation proxy, as degradation has been shown to materially affect energy arbitrage profitability \citep{wankmuller2017impact}. 
\subsubsection*{Manhattan EMS Maximum Coverage}

The empirical setting is an hourly ZIP-level EMS demand prediction and coverage problem in Manhattan. The pipeline counts all EMS rows by hour for the citywide lag feature. It then filters to the Manhattan borough, extracts five-digit ZIP codes, and counts incidents by hour and ZIP. The EMS incident data are sourced from NYC Open Data \citep{ems2016}.

The ZIP geography is built from modified ZIP code tabulation area (MODZCTA) fields from geodatafiles available online \citep{modzcta2020}. Alias ZIPs are mapped to a canonical MODZCTA ZIP and incident counts are summed after mapping. The dataset contains 45 ZIP zones and 5,832 hourly observations from January 1, 2025 through August 31, 2025, for 262,440 ZIP-hour rows. Missing incident counts in the dense ZIP-hour grid are set to zero. The prediction target is the vector of next-hour EMS incident counts across the 45 ZIP zones. The target sum over the output window is 252,665 incidents, corresponding to approximately 43.3 incidents per hour citywide. The processed shared features are hour, day of the week, temperature, precipitation, citywide incidents lagged by one hour, ZIP-level incidents lagged by one hour, mean neighboring incidents lagged by one hour, and the zone-hour historical baseline, calculated over the past 8 weeks for each row. The model uses 140 covariates grouped into eight players: hour of day, day of week, temperature, precipitation, citywide lag-1 EMS demand, ZIP-level lag-1 EMS demand, neighboring-zone lag-1 demand, and the zone-hour historical baseline.

The training period is January 1, 2025 through July 31, 2025, with 5,088 training hours. The holdout period is August 1, 2025 through August 31, 2025, with 744 holdout hours and 33,480 ZIP-hour predictions. The explanation sample contains 100 holdout hours sampled without replacement with random seed 0. The EMS background set contains 100 training hours sampled without replacement with random seed 0. For each coalition, included feature-group columns are set to the explained hour's values, excluded columns remain at sampled background values, XGBoost predictions are averaged over background rows, and negative predictions are truncated to zero. For each holdout hour, the predicted 45-dimensional demand vector is passed to a downstream maximum coverage problem that selects a limited number of EMS staging zones. A selected staging zone covers a demand zone if the distance between their ZIP centroids is within a specified coverage radius. We evaluate three coverage radii, 1, 2, and 3 km, and three facility budgets, 3, 5, and 8 staging locations.
\section*{Appendix B. Additional Figures \& Results}
\begin{figure}[ht]
    \centering
    \includegraphics[width=0.83\linewidth, trim={0 20 0 18}, clip]{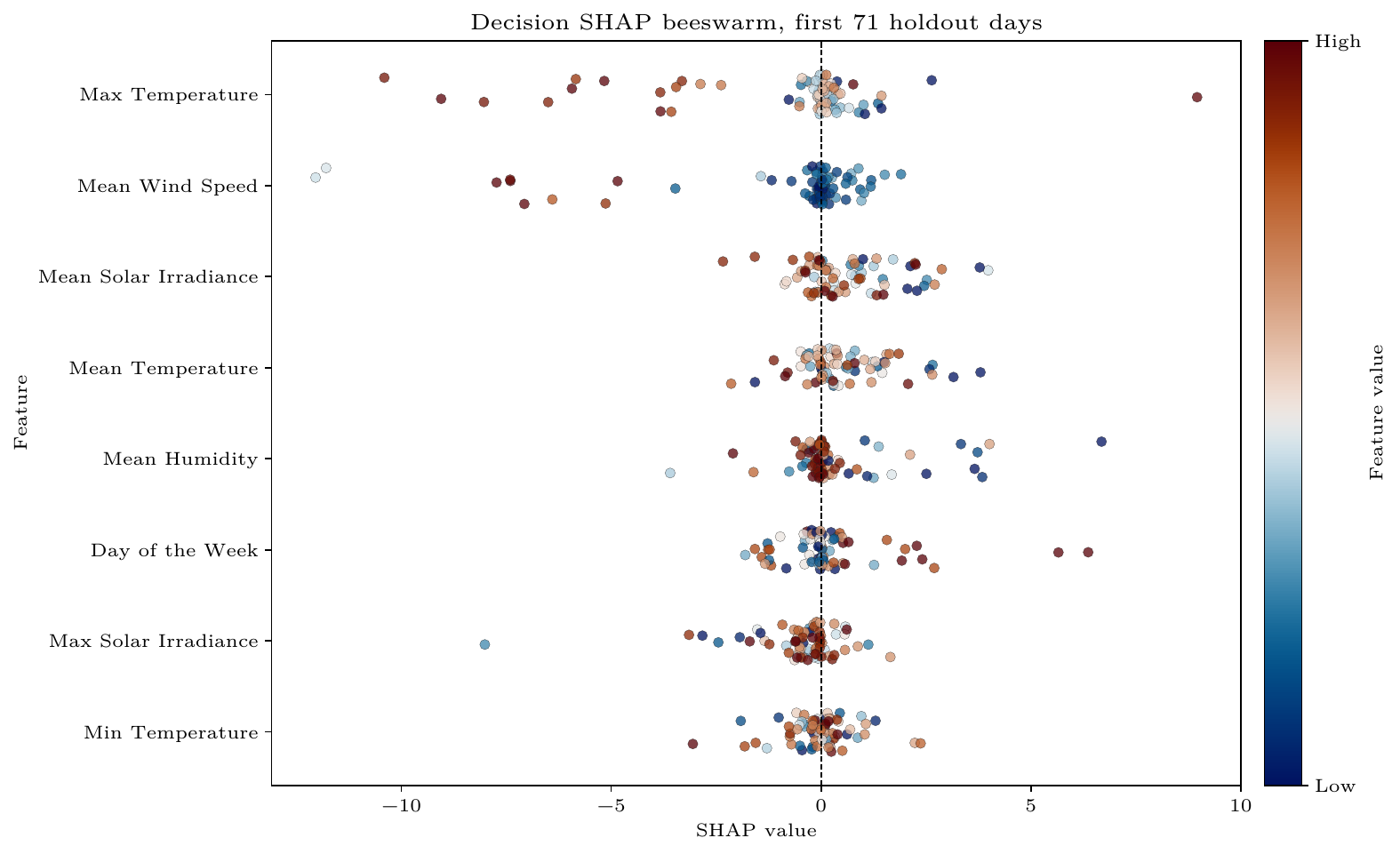}
    \caption{A beeswarm plot for the post-InfoDVA values on the test dataset. Each point corresponds to a single day's SHAP value for the corresponding feature, with the color denoting the relative value of the feature at that point. We see that the mean wind speed has the lowest post-InfoDVA values with very few positive instances.}
    \label{fig:caiso-beeswarm}
\end{figure}

\begin{figure}[ht]
    \centering
    \includegraphics[width=\linewidth]{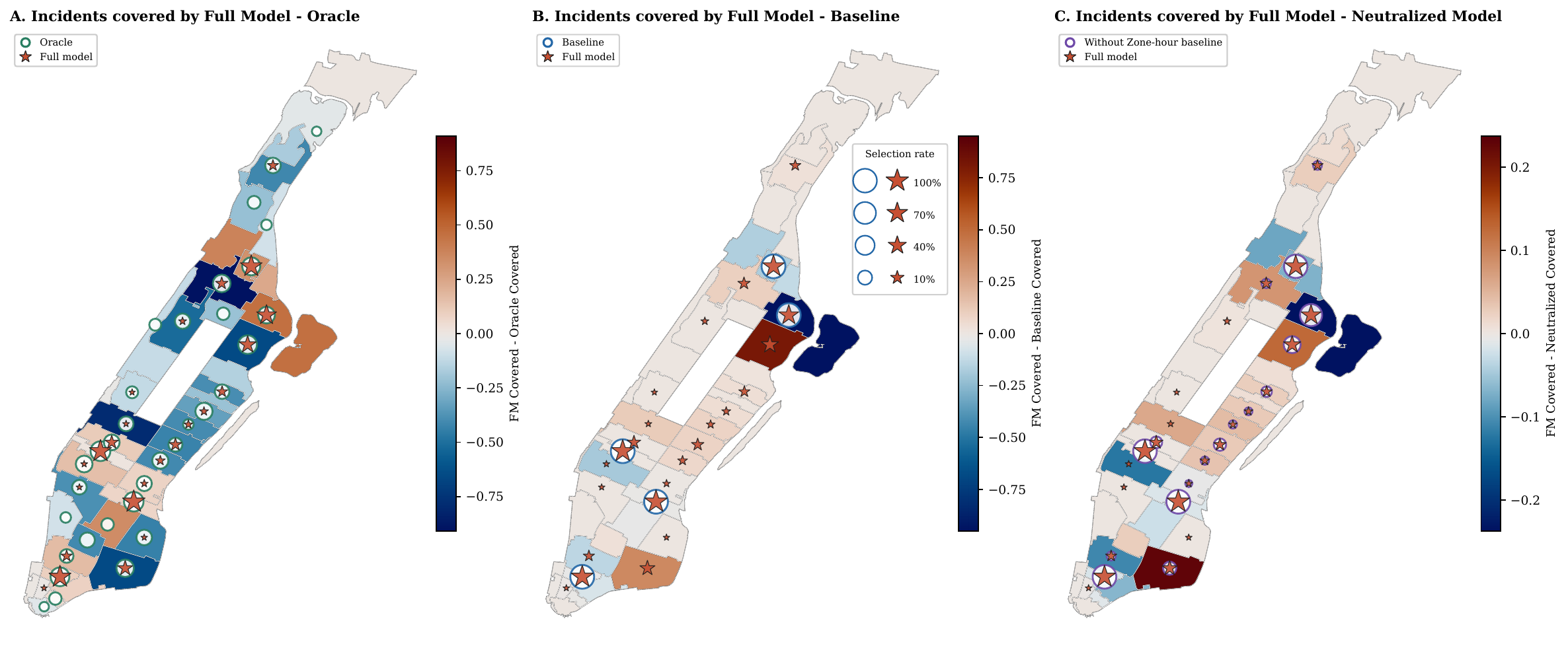}
    \caption{Spatial distribution of decision value difference between different predictors for the $p=5, \tau = 1$ design. In (A), we compare the full model with the oracle, which has access to perfect information. In (B), we compare the full model $\hat y_N$ with the uninformed baseline $\hat y_\emptyset$. In (C), we compare the full model $\hat y_N$ with $\hat y_{N \backslash \{\mathrm{ZH}\}}$, where $\mathrm{ZH}$ is the zone-hour baseline feature. The points show the frequency that a zone was chosen as a staging area by each predictor. This shows how the perfect information can change the policy (A), how input features can induce changes in the downstream policy (B), and how the most influential feature, ZH, can change the policy when introduced to the model.}
    \label{fig:ems-spatial-decisionmaking}
\end{figure}
\begin{sidewaysfigure}
    \centering
    \includegraphics[width=1\linewidth, trim={0 0 0 18}, clip]{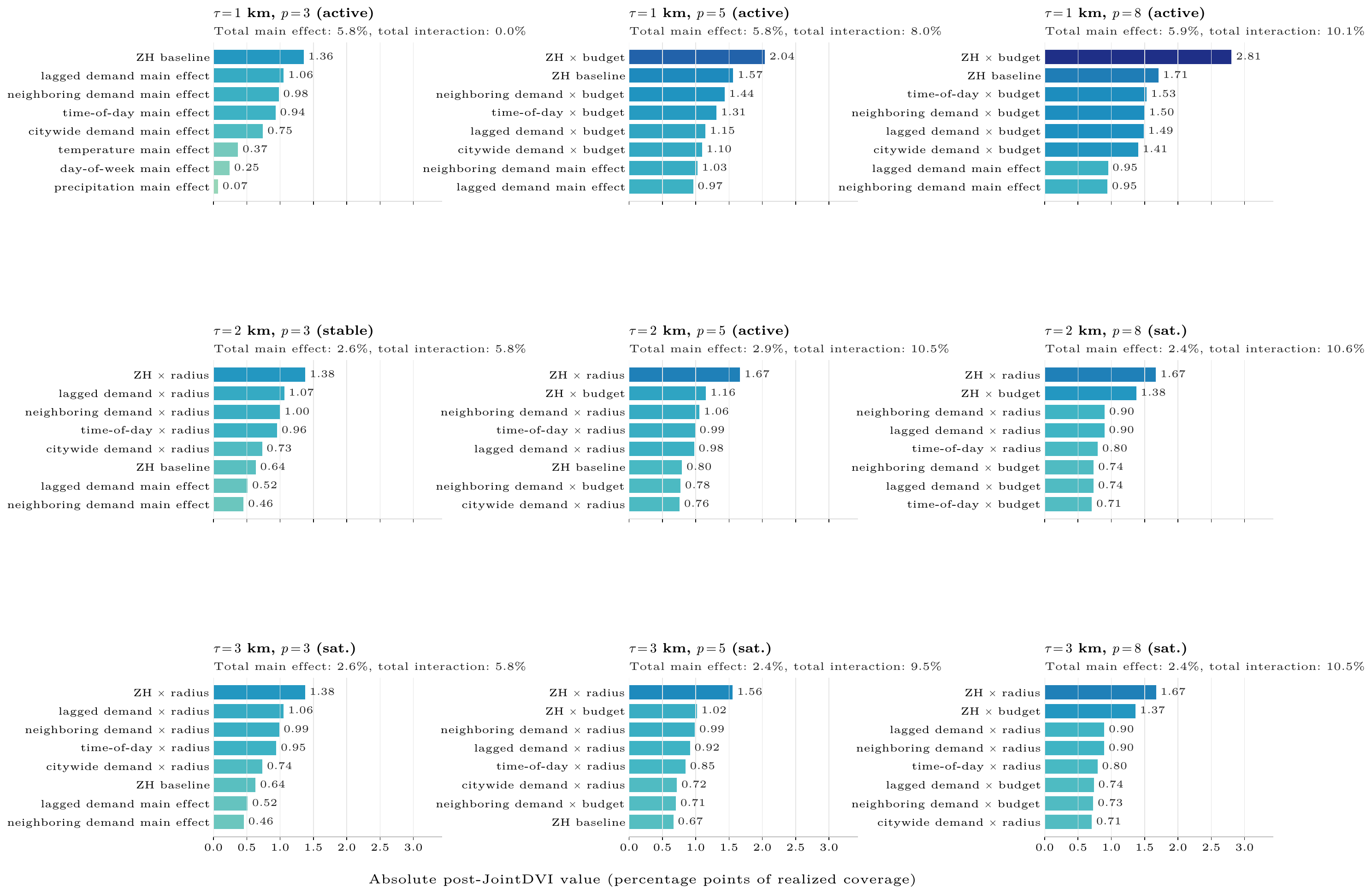}
    \caption{Each panel reports the largest post-JointDVI main or interaction terms for one budget-radius regime, with values expressed in mean percentage points of realized coverage for each design. The figure shows that the zone-hour baseline feature group is consistently the largest contributor, but that its contribution appears primarily through interactions with budget in low-radius regimes and with radius in higher-radius regimes.}
    \label{fig:ems-joint-dvi-detailed}
\end{sidewaysfigure}
\clearpage
\subsection*{SHAP Approximation Methods}
We implement two popular approximation methods for recovering SHAP values and apply them to DVA: Kernel SHAP and Permutation SHAP, which we introduced in \cref{subsec:compute-complexity}. The goal of this analysis is to quantify the performance of these methods in estimating the exact SHAP values with substantially less compute time. For each method, we test 5 different budgets: $\{16,32,64,128,192\}$, where 256 are the number of coalitions the exact DVA attribution method must evaluate. For each method-budget combination, we test across 50 different seeds to account for variety in sampling. We report the Kendall tau similarity of the global rankings produced and the agreement \% of the global top-1 feature with the exact InfoDVA attribution. We also compute a normalized MAE (NMAE) of the InfoDVA values from the exact values, to test how well the exact values are approximated by each method. Formally, let $T$ be the number of explained EMS hours, $d=8$ the number of feature groups, $\phi_{t i}^\mathrm{exact}$ the exact InfoDVA value for hour $t$ and feature $i$, and $\hat{\phi}_{i t}^{(r,B)}$ the approximate InfoDVA value for random seed $r$ and budget $B$. We compute NMAE as
\begin{equation*}
    \mathrm{NMAE}^{(r,B)} = \frac{\sum_{t = 1}^T \sum_{i = 1}^d \lvert\hat{\phi}_{i t}^{(r,B)} - \phi_{t i}^\mathrm{exact}\rvert}{\sum_{t = 1}^T \sum_{i = 1}^d \lvert\phi_{t i}^\mathrm{exact}\rvert}.
\end{equation*}
All results are accompanied by 95\% confidence intervals across 10,000 bootstrap sets for the 50 seeded runs.
We report the results in \cref{tab:ems-approx-decision-shap}.
\begin{table}[ht]
\centering
\caption{Approximation quality for EMS InfoDVA in the decision-active configuration $(p=8,\tau=1)$. Exact DVA enumerates all $2^8=256$ coalitions per explained hour. Kernel budgets (B) represent sampled orderings, while permutation budgets (B) denote sampled feature orderings. Oracle calls and runtimes (in seconds) use cached unique decision-value evaluations and are averaged over 50 approximation runs per method per budget, and reported with 95\% bootstrap CI.}
\label{tab:ems-approx-decision-shap-full}
\scriptsize
\begin{tabular}{@{}lrr S[table-format=2.2]@{\,}l l l r@{}}
\toprule
Estimator & B & Calls & \multicolumn{2}{c}{Time (\% exact)} & NMAE & Kendall tau & Top-1 \\
\midrule
Exact & 256 & 25,600 & 25.44 & (100.0\%) & --- & --- & --- \\
Kernel & 16  & 1,800  & 1.79  & (7.0\%)  & 1.441 [1.413, 1.472] & 0.749 [0.717, 0.779] & 76\% \\
Kernel & 32  & 3,400  & 3.38  & (13.3\%) & 0.935 [0.919, 0.954] & 0.851 [0.833, 0.870] & 90\% \\
Kernel & 64  & 6,600  & 6.56  & (25.8\%) & 0.655 [0.644, 0.666] & 0.887 [0.871, 0.903] & 92\% \\
Kernel & 128 & 13,000 & 12.92 & (50.8\%) & 0.400 [0.395, 0.404] & 0.946 [0.933, 0.959] & 100\% \\
Kernel & 192 & 19,400 & 19.28 & (75.8\%) & 0.222 [0.217, 0.228] & 0.966 [0.956, 0.976] & 100\% \\
Permutation & 16  & 8,324  & 8.27  & (32.5\%) & 0.473 [0.468, 0.480] & 0.929 [0.916, 0.940] & 96\% \\
Permutation & 32  & 13,118 & 13.03 & (51.2\%) & 0.337 [0.332, 0.341] & 0.940 [0.933, 0.949] & 100\% \\
Permutation & 64  & 18,735 & 18.62 & (73.2\%) & 0.238 [0.235, 0.241] & 0.967 [0.957, 0.977] & 100\% \\
Permutation & 128 & 23,324 & 23.19 & (91.1\%) & 0.169 [0.167, 0.172] & 0.959 [0.949, 0.969] & 100\% \\
Permutation & 192 & 24,805 & 24.67 & (97.0\%) & 0.139 [0.138, 0.141] & 0.963 [0.953, 0.973] & 100\% \\
\bottomrule
\end{tabular}
\end{table}

Our results in \cref{tab:ems-approx-decision-shap-full} show that approximation SHAP is an effective solution for mitigating the computational complexity of DVA. Permutation SHAP is shown to be very strong at rank recovery. Even 16 sampled orderings reaches a Kendall tau of 0.929 and 96\% top-1 agreement. Kendall SHAP on the other hand converges smoothly with budget. While weaker at smaller budgets, it recovers the global ranking successfully at around 128-192 sampled coalitions. Furthermore, at comparable cached oracle call budgets, both methods are highly reliable. At around 18k-19k calls, both methods reach a Kendall tau of $\approx 0.966$ with NMAE around 0.22-0.24. Some good choices for approximations are Permutation-16, with a strong ranking recovery at only 32.5\% the runtime, Kernel-128, with 100\% top-1 recovery and a Kendall tau of 0.946 at half the total oracle calls, and Permutation-64 or Kernel-192, with near-exact global rank recovery at 73-76\% the total time budget.
\end{document}